\pdfoutput=1
\documentclass[conference]{IEEEtran}

\usepackage{cite}
\usepackage{amsmath,amssymb,amsfonts}
\usepackage{graphicx}
\usepackage{booktabs}
\usepackage{multirow}
\usepackage{hyperref}
\hypersetup{hidelinks}

\begin{document}

\title{CommDreamer: Radio-Geometric World Models for Mapless Cellular-Connected UAV Trajectory Planning}

\author{
	\IEEEauthorblockN{Yuqi Ping\IEEEauthorrefmark{1}, Tianhao Liang\IEEEauthorrefmark{1}, and Tingting Zhang\IEEEauthorrefmark{1}\IEEEauthorrefmark{3}}
	\IEEEauthorblockA{\IEEEauthorrefmark{1}Guangdong Provincial Key Laboratory of Space-Aerial Networking and Intelligent Sensing,\\
	Harbin Institute of Technology, Shenzhen, China}
	\IEEEauthorblockA{\IEEEauthorrefmark{3}Peng Cheng Laboratory (PCL), Shenzhen, China}
}

\maketitle
\bstctlcite{IEEEexample:BSTcontrol}

\begin{abstract}
Reliable cellular connectivity and collision-free flight are essential for unmanned aerial vehicle (UAV) operations in low-altitude urban canyons, where dense 3D buildings obstruct flight clearance, shadow serving base stations, and create strong co-channel interference at street intersections. Existing planners either require prior 3D city meshes and channel knowledge maps unavailable onboard, or rely on model-free reinforcement learning with low sample efficiency and poor cross-city generalization. We propose \textit{CommDreamer}, a mapless radio-geometric world model that fuses raw onboard 3D LiDAR scans, multicell cellular reports, and shift-invariant relative kinematics to jointly infer local obstacle clearance and directional signal-to-interference-plus-noise ratio (SINR) dynamics. Using a categorical recurrent state-space model, \textit{CommDreamer} compresses multimodal measurement histories into a coordinate-invariant latent belief, enabling the UAV to foresee deep shadow zones and train a lightweight online policy purely via multi-step latent imagination without prior maps or antenna patterns. Zero-shot evaluations across unseen urban morphologies in our open-source ray-tracing simulator, \textit{AeroCity-RT}, show that \textit{CommDreamer} converges within $150$ episodes where model-free reinforcement learning fails to complete flights, achieving $100.0\%$ mission success with zero collisions while lifting median route SINR to $8.02\,\mathrm{dB}$ and cutting effective SINR outage by over $50\%$ relative to radio-blind world-model and reactive LiDAR baselines.
\end{abstract}


\section{Introduction}

Rapid growth of the low-altitude economy has accelerated urban deployments of unmanned aerial vehicles (UAVs) for parcel delivery, infrastructure inspection, and emergency rescue~\cite{ping2026wcm,ping2026nsaai}. Safe execution of such missions requires reliable command, control, and payload links between the UAV and ground cellular base stations (BSs)~\cite{zeng2019accessing}. Unlike flights at high altitude with dominant line-of-sight (LoS) propagation, urban airspace below rooftop height experiences coupled collision risks and abrupt coverage degradation: dense 3D buildings obstruct flight clearance, cast deep non-LoS (NLoS) shadow fading on serving BSs, and expose UAVs at street intersections to strong LoS co-channel interference from neighboring sectors, causing severe signal-to-interference-plus-noise ratio (SINR) outages~\cite{ping2025gcs,ping2026stl_urllc}. Safe and connectivity-aware trajectory planning in unfamiliar urban canyons is therefore essential.

Early trajectory planning relied on mathematical optimization under statistical path loss or probabilistic LoS models~\cite{zhang2019cellular,ping2025gcs}, which average out local building geometry and miss sharp shadowing transitions. Subsequent studies adopted channel knowledge maps (CKMs)~\cite{zeng2021ckm} and ray tracing simulators such as Sionna RT~\cite{hoydis2023sionna_rt} for site specific channel modeling, yet both offline trajectory optimization~\cite{ping2026llm_stl} and CKM aided planning depend on prior information unavailable onboard during flight, including global 3D city meshes, full area radio maps, and exact BS antenna patterns. Although deep reinforcement learning (DRL) relaxes explicit channel priors in communication-aware trajectory planning~\cite{challita2018interference,zeng2021simultaneous}, existing DRL planners still suffer from poor generalization across cities and low sample efficiency: policies conditioned on global coordinates over a fixed BS layout memorize a city specific SINR lookup map that fails in unseen cities~\cite{challita2018interference}, whereas model-free value and actor-critic planners~\cite{zeng2021simultaneous} map instantaneous observations directly to actions without a predictive model of blockage and interference evolution, requiring extensive interactions and failing to foresee abrupt corner outages~\cite{hao2026look}.

To address these challenges, we propose \textit{CommDreamer}, a radio-geometric world model for mapless cellular-connected UAV trajectory planning in unfamiliar urban canyons. Our core physical insight is that the omnidirectional 3D building facades captured by raw onboard LiDAR point clouds are the exact scatterers governing directional LoS blockage and intercell interference shielding, inherently coupling visible street geometry with multicell radio propagation. Inspired by DreamerV3~\cite{hafner2023dreamerv3}, which decouples dynamics modeling from policy learning via imagined rollouts, \textit{CommDreamer} compresses local LiDAR and cellular measurement histories into a compact latent belief invariant to global coordinates, eliminating explicit 3D maps and global CKMs. Predicting future trajectories in latent space enables the UAV to foresee upcoming shadow zones and interference surges, bypassing obstacle traps and coverage holes with a lightweight online policy. Our main contributions are threefold:
\begin{enumerate}
	\item We formulate mapless cellular-connected UAV trajectory planning as a partially observable Markov decision process (POMDP) invariant to coordinate shifts, driven solely by raw onboard 3D LiDAR point clouds, multicell RSRP and SINR reports, and relative BS and goal kinematics without prior 3D maps or BS antenna patterns.
	\item We develop a multimodal radio-geometric world model fusing spherical LiDAR projections with symlog encoded cellular and kinematic features inside a categorical recurrent state space model (RSSM), jointly inferring local obstacle clearance, directional LoS/NLoS shadowing, and multicell interference dynamics for multi-horizon latent trajectory optimization.
	\item We build an open-source 3D urban wireless simulator, \textit{AeroCity-RT}\footnote{\url{https://github.com/ppppyq/AeroCity-RT}}, coupling procedural city generation and 3GPP multicell deployments with Sionna RT ray tracing~\cite{hoydis2023sionna_rt}. Zero-shot evaluations across unseen urban morphologies show that \textit{CommDreamer} converges within $150$ episodes where model-free DRL fails to complete flights, achieves $100.0\%$ mission success with zero collisions, boosts mean route SINR by $1.59\text{--}1.78\,\mathrm{dB}$ to $7.69\,\mathrm{dB}$ and median SINR by $3.31\text{--}3.64\,\mathrm{dB}$ to $8.02\,\mathrm{dB}$, and cuts effective link outage by $50.1\text{--}51.5\%$ down to $12.0\%$ over radio-blind world-model and reactive LiDAR baselines.
\end{enumerate}

\section{System Model and Problem Formulation}

This section introduces the 3D urban multicell channel model, UAV flight dynamics and obstacle avoidance constraint, onboard multimodal sensing model, and trajectory planning problem formulation.

\subsection{3D Urban Environment and Multi-Cell Channel Model}

Consider a cellular-connected UAV performing a point-to-point flight mission in a 3D urban region $\mathcal{W} \subset \mathbb{R}^3$ over $T$ discrete control slots $\mathcal{T} \triangleq \{0, 1, \dots, T-1\}$ of duration $\Delta t$, where $T$ denotes the variable mission completion step. The region contains $N_{\mathrm{b}}$ buildings, with $\mathcal{B}_b$ denoting the volume occupied by the $b$-th building. The obstacle set and obstacle-free airspace are defined as $\mathcal{B} = \bigcup_{b=1}^{N_{\mathrm{b}}} \mathcal{B}_b$ and $\mathcal{W}_{\mathrm{free}} = \mathcal{W} \setminus \mathcal{B}$, respectively, and $M$ directional ground BS sectors, indexed by $\mathcal{M} \triangleq \{1, \dots, M\}$, operate across $\mathcal{W}$ under universal frequency reuse. From existing urban cellular registries and mission specifications, the BS antenna locations $\{\mathbf{p}_{\mathrm{BS},m}\}_{m=1}^M$ and mission destination $\mathbf{p}_{\mathrm{G}} \in \mathcal{W}_{\mathrm{free}}$ are known to the UAV before flight, whereas the 3D building layout $\mathcal{B}$, BS antenna radiation patterns, and spatial channel distribution across $\mathcal{W}$ are unavailable to the UAV in unfamiliar cities.

At step $t$, given UAV position $\mathbf{p}_t \in \mathcal{W}_{\mathrm{free}}$, BS transmit power $P_{\mathrm{tx},m}$, reference power $P_{\mathrm{ref}}$, and air-to-ground channel power gain $G_m(\mathbf{p}_t) > 0$ from sector $m$, the received reference signal power and corresponding RSRP in dBm are
\begin{equation}
	P_{m,t}^{\mathrm{rx}} = P_{\mathrm{tx},m} G_m(\mathbf{p}_t),
	\quad
	R_{m,t} = 10\log_{10}\!\left(\frac{P_{m,t}^{\mathrm{rx}}}{P_{\mathrm{ref}}}\right)\!.
	\label{eq:spawc-rsrp}
\end{equation}
Following 3GPP NR measurement reporting procedures~\cite{zeng2019accessing,zhang2019cellular}, the onboard cellular modem detects the $M_{\mathrm{obs}}$ strongest BS sectors ordered in descending RSRP as $(m_1, \dots, m_{M_{\mathrm{obs}}})$, where $m_1 = \arg\max_{m \in \mathcal{M}} P_{m,t}^{\mathrm{rx}}$ is the serving sector, while all remaining sectors generate co-channel interference scaled by traffic load $\rho_{\mathrm{load}} \in (0, 1]$. Given receiver noise power $\sigma_{\mathrm{n}}^2$ with noise floor $R_{\mathrm{n}} = 10\log_{10}(\sigma_{\mathrm{n}}^2 / P_{\mathrm{ref}})$ in dBm, the post-association SINR in dB reported by the modem together with $\{R_{m_k,t}\}_{k=1}^{M_{\mathrm{obs}}}$ is
\begin{equation}
	\gamma_t = 10\log_{10}\!\left( \frac{P_{m_1,t}^{\mathrm{rx}}}{\sigma_{\mathrm{n}}^2 + \rho_{\mathrm{load}} \sum_{j \in \mathcal{M} \setminus \{m_1\}} P_{j,t}^{\mathrm{rx}}} \right)\!.
	\label{eq:spawc-sinr}
\end{equation}
Given a minimum SINR threshold $\gamma_{\min}$ for reliable connectivity, the binary communication outage indicator at step $t$ is
\begin{equation}
	O_t = \mathbb{I}(\gamma_t < \gamma_{\min}) \in \{0, 1\},
	\label{eq:spawc-outage-def}
\end{equation}
where $\mathbb{I}(\cdot)$ denotes the indicator function.

\subsection{UAV Flight Dynamics and Safety Constraint}

Following low-altitude UTM altitude-layering guidelines~\cite{faa2020utm}, the UAV flies at a constant cruise altitude $H_{\mathrm{uav}}$. Starting from $\mathbf{p}_0 \in \mathcal{W}_{\mathrm{free}}$, it completes the mission at the first step $T$ satisfying $\|\mathbf{p}_T - \mathbf{p}_{\mathrm{G}}\|_2 \le d_{\mathrm{goal}}$. At each step $t$, the planner issues a horizontal acceleration command $\mathbf{a}_t \in \mathcal{A} \triangleq \{\mathbf{a} \in \mathbb{R}^2 : \|\mathbf{a}\|_2 \le a_{\max}\}$ under a maximum horizontal speed $v_{\max}$. By quadrotor differential flatness, $\mathbf{a}_t$ and aerodynamic drag determine the desired body tilt and velocity reference tracked by the inner-loop attitude-velocity controller. Denoting by $\mathbf{x}_t^{\mathrm{uav}}$ the UAV state comprising position $\mathbf{p}_t = [x_t, y_t, H_{\mathrm{uav}}]^\top$, horizontal velocity $\mathbf{v}_t \in \mathbb{R}^2$, and 3D attitude $\mathbf{R}_t \in \mathrm{SO}(3)$ with roll, pitch, and yaw $(\phi_t, \theta_t, \psi_t)$, the closed-loop flight dynamics are given by
\begin{equation}
	\mathbf{x}_{t+1}^{\mathrm{uav}} = \mathcal{F}_{\mathrm{dyn}}\big(\mathbf{x}_t^{\mathrm{uav}}, \mathbf{a}_t\big).
	\label{eq:spawc-dyn}
\end{equation}
Throughout the flight, the UAV must remain strictly outside all 3D buildings:
\begin{equation}
	d_{\mathrm{obs},t} \triangleq \inf_{\mathbf{q} \in \mathcal{B}} \|\mathbf{p}_t - \mathbf{q}\|_2 > 0,
	\quad \forall t \in \mathcal{T} \cup \{T\}.
	\label{eq:spawc-collision}
\end{equation}

\subsection{Onboard Multimodal Perception}

In the absence of prior 3D building layouts $\mathcal{B}$ and channel maps $\{G_m(\mathbf{p})\}_{m \in \mathcal{M}}$, the UAV relies on real-time onboard sensing for trajectory planning. Specifically, an onboard body-mounted 3D LiDAR scans $N_{\mathrm{az}}$ azimuth and $N_{\mathrm{el}}$ elevation angles within a maximum range $R_{\mathrm{L}}$. Because maneuvers induce body tilt $(\phi_t, \theta_t)$, each raw point measured in the body frame $\mathcal{F}_{\mathrm{B}}$ is rotated via the IMU-measured attitude matrix $\mathbf{R}_{\mathrm{B}}^{\mathrm{L}}(\phi_t, \theta_t)$ into a gravity-aligned UAV-centered local horizontal frame $\mathcal{F}_{\mathrm{L}}$, forming a spherical range image
\begin{equation}
	\mathbf{L}_t = \mathcal{F}_{\mathrm{LiDAR}}\big(\mathbf{x}_t^{\mathrm{uav}}; \mathcal{B}\big) \in [0, R_{\mathrm{L}}]^{N_{\mathrm{el}} \times N_{\mathrm{az}}}.
	\label{eq:spawc-lidar}
\end{equation}
Meanwhile, the onboard navigation system and cellular modem provide $\mathbf{p}_t$, $\mathbf{v}_t$, and $(\gamma_t, \{R_{m_k,t}\}_{k=1}^{M_{\mathrm{obs}}})$. To avoid overfitting to global coordinates and align spatial scales with $\mathbf{L}_t$, the relative displacements to the detected BS sectors and destination are normalized by $R_{\mathrm{L}}$:
\begin{equation}
	\mathbf{r}_{k,t}^{\mathrm{BS}} = \frac{\mathbf{p}_{\mathrm{BS},m_k} - \mathbf{p}_t}{R_{\mathrm{L}}},
	\quad
	\mathbf{r}_t^{\mathrm{G}} = \frac{\mathbf{p}_{\mathrm{G}} - \mathbf{p}_t}{R_{\mathrm{L}}}.
	\label{eq:spawc-rel-vec}
\end{equation}
Combining the serving-sector SNR $R_{m_1,t} - R_{\mathrm{n}}$, neighbor RSRP margins $R_{m_1,t} - R_{m_k,t}$ for $k = 2, \dots, M_{\mathrm{obs}}$, and relative displacements of the serving and strongest interfering sectors $m_1, m_2$, the cellular radio and ego-kinematic vectors are formed as
\begin{equation}
	\begin{aligned}
		\mathbf{s}_t^{\mathrm{radio}} = \Big[ &\gamma_t, \; R_{m_1,t} - R_{\mathrm{n}}, \; \{R_{m_1,t} - R_{m_k,t}\}_{k=2}^{M_{\mathrm{obs}}}, \\
		&\mathbf{r}_{1,t}^{\mathrm{BS}\top}, \; \|\mathbf{r}_{1,t}^{\mathrm{BS}}\|_2, \; \mathbf{r}_{2,t}^{\mathrm{BS}\top}, \; \|\mathbf{r}_{2,t}^{\mathrm{BS}}\|_2 \Big]^\top\!,
	\end{aligned}
	\label{eq:spawc-radio-vec}
\end{equation}
\begin{equation}
	\mathbf{x}_t^{\mathrm{ego}} = \big[ \mathbf{r}_t^{\mathrm{G}\top}, \; \|\mathbf{r}_t^{\mathrm{G}}\|_2, \; \mathbf{v}_t^\top / v_{\max} \big]^\top\!.
	\label{eq:spawc-ego-vec}
\end{equation}

\begin{figure*}[!t]
	\centering
	\includegraphics[trim=12bp 10bp 12bp 8bp,clip,width=0.86\textwidth]{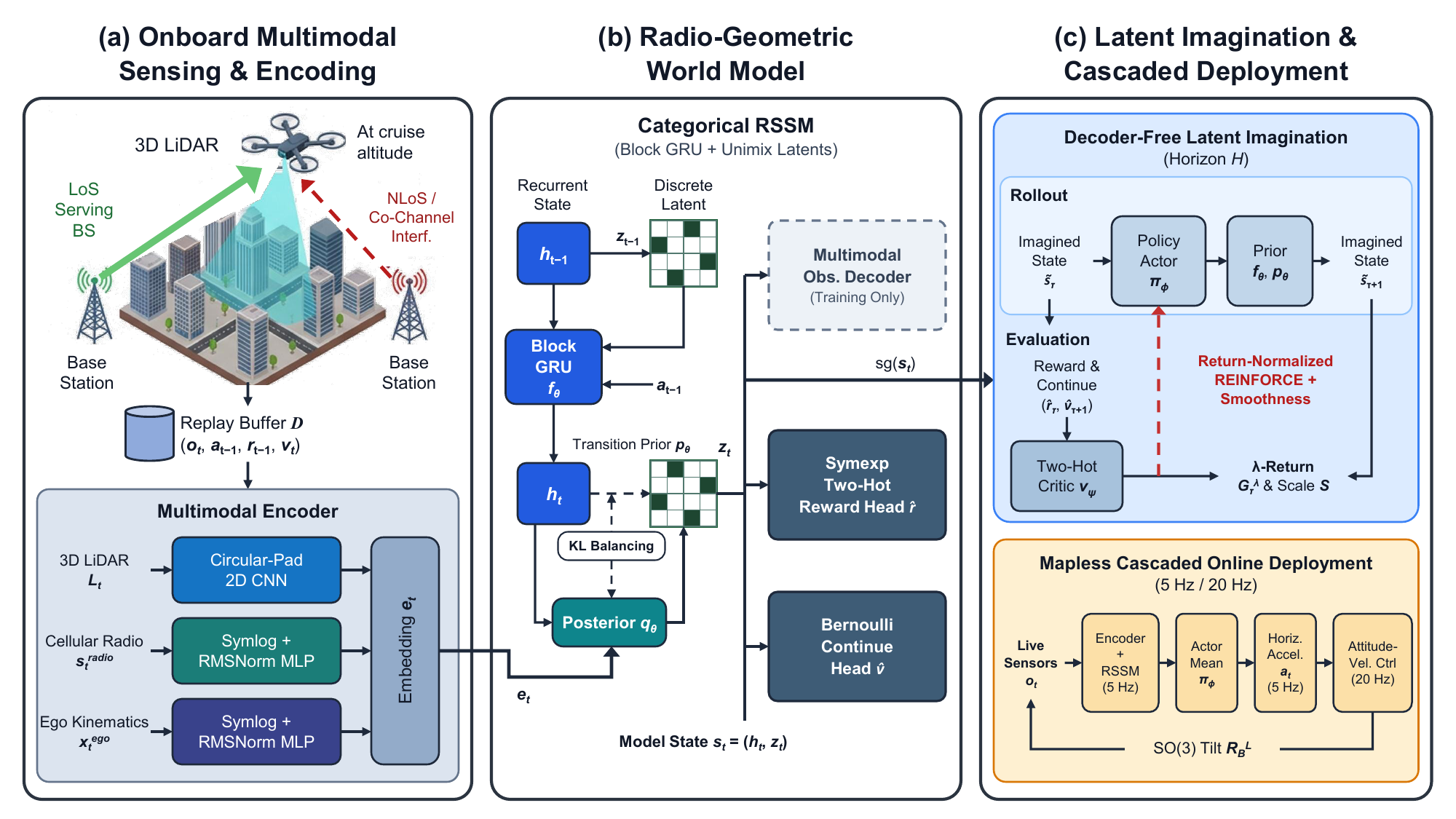}
	\caption{Overall architecture of \textit{CommDreamer}: (a) coordinate-invariant onboard multimodal sensing fusing spherical 3D LiDAR scans $\mathbf{L}_t$, cellular reports $\mathbf{s}_t^{\mathrm{radio}}$, and ego-relative kinematics $\mathbf{x}_t^{\mathrm{ego}}$, (b) radio-geometric categorical RSSM trained with observation, symexp two-hot reward, and Bernoulli continuation heads, and (c) decoder-free latent imagination via bootstrapped $\lambda$-returns and return-normalized REINFORCE with mapless cascaded online deployment coupling a 5~Hz outer policy and a 20~Hz inner attitude-velocity controller.}
	\label{fig:spawc-framework}
	\vspace{-0.20cm}
\end{figure*}

\subsection{Problem Formulation}

The mission aims to reach the destination as rapidly as possible while minimizing communication outage time and control effort, subject to flight dynamics and collision avoidance:
\begin{align*}
	(\mathcal{P}_1): \min_{T, \{\mathbf{a}_t\}} \; &\sum_{t=0}^{T-1} \!\Big( \Delta t + w_{\mathrm{out}} \Delta t \, O_{t+1} + w_a \|\mathbf{a}_t\|_2^2 \Big) \\
	\mathrm{s.t.} \quad &\eqref{eq:spawc-dyn}, \; \|\mathbf{a}_t\|_2 \le a_{\max}, \; \|\mathbf{v}_{t+1}\|_2 \le v_{\max}, \; \forall t \in \mathcal{T}, \\
	&\eqref{eq:spawc-collision}, \; \|\mathbf{p}_T - \mathbf{p}_{\mathrm{G}}\|_2 \le d_{\mathrm{goal}},
\end{align*}
where $w_{\mathrm{out}}, w_a > 0$ weight the outage duration and acceleration effort relative to flight time. Classical offline trajectory optimization~\cite{zhang2019cellular,ping2025gcs} cannot solve $(\mathcal{P}_1)$ for two reasons. First, since $\mathcal{B}$ and $\{G_m(\mathbf{p})\}_{m \in \mathcal{M}}$ are unknown, neither future obstacle clearance nor future SINR can be evaluated offline, requiring closed-loop decisions from local measurements. Second, $(\mathcal{P}_1)$ involves a variable horizon $T$, a non-convex obstacle field, and a discontinuous outage indicator $O_{t+1}$ whose spatial gradient vanishes almost everywhere. We therefore recast $(\mathcal{P}_1)$ into a POMDP and solve it via the \textit{CommDreamer} radio-geometric world model.

\section{CommDreamer: POMDP Reformulation and Radio-Geometric World Model}

As illustrated in Fig.~\ref{fig:spawc-framework}, we first reformulate $(\mathcal{P}_1)$ into a POMDP in Section~III-A, and then solve it via \textit{CommDreamer} following the DreamerV3 paradigm~\cite{hafner2023dreamerv3} through multimodal belief-state world modeling in Section~III-B and latent-imagination policy optimization in Section~III-C.

\subsection{POMDP Reformulation}

We cast $(\mathcal{P}_1)$ as a POMDP $\mathcal{G} = (\mathcal{S}, \allowbreak \mathcal{A}, \allowbreak \mathcal{P}, \allowbreak \mathcal{O}, \allowbreak \mathcal{Z}, r, \eta)$ with discount factor $\eta \in (0, 1)$, specified by three components.

\textbf{State and Transition Dynamics:}
The environment state $\mathbf{s}_t^{\mathrm{env}} = (\mathbf{x}_t^{\mathrm{uav}}, \allowbreak \mathbf{p}_{\mathrm{G}}, \allowbreak \{\mathbf{p}_{\mathrm{BS},m}\}_{m=1}^M, \allowbreak \mathcal{B}, \allowbreak \{G_m\}_{m \in \mathcal{M}}) \in \mathcal{S}$ gathers the UAV state, mission coordinates, building layout $\mathcal{B}$, and channel fields $\{G_m\}_{m \in \mathcal{M}}$. Under horizontal acceleration $\mathbf{a}_t \in \mathcal{A}$, the transition kernel $\mathcal{P}(\mathbf{s}_{t+1}^{\mathrm{env}} \mid \mathbf{s}_t^{\mathrm{env}}, \mathbf{a}_t)$ advances $\mathbf{x}_t^{\mathrm{uav}}$ via~\eqref{eq:spawc-dyn} while keeping $\mathcal{B}$ and $\{G_m\}_{m \in \mathcal{M}}$ static. The continuation flag $\nu_{t+1} \in \{0, 1\}$ turns $0$ upon goal arrival $\|\mathbf{p}_{t+1} - \mathbf{p}_{\mathrm{G}}\|_2 \le d_{\mathrm{goal}}$ or collision $d_{\mathrm{obs},t+1} \le 0$, terminating the trajectory at step $T$.

\textbf{Observation Space:}
Induced by the onboard multimodal sensors in Section~II-C, the observation kernel $\mathcal{Z}(\mathbf{o}_t \mid \mathbf{s}_t^{\mathrm{env}})$ emits the shift-invariant, prior-free observation tuple
\begin{equation}
	\mathbf{o}_t = \big( \mathbf{L}_t, \; \mathbf{s}_t^{\mathrm{radio}}, \; \mathbf{x}_t^{\mathrm{ego}} \big) \in \mathcal{O}.
	\label{eq:spawc-obs}
\end{equation}

\textbf{Reward Function and Policy Objective:}
To reflect the objective and constraints of $(\mathcal{P}_1)$, the step reward $r_t = r(\mathbf{s}_t^{\mathrm{env}}, \mathbf{a}_t, \mathbf{s}_{t+1}^{\mathrm{env}})$ with weights $w_{\mathrm{sinr}}, w_{\mathrm{obs}} > 0$ is
\begin{equation}
	r_t = r_t^{\mathrm{nav}} + w_{\mathrm{sinr}} \, r_t^{\mathrm{sinr}} + w_{\mathrm{obs}} \, r_t^{\mathrm{obs}}.
	\label{eq:spawc-reward}
\end{equation}
Letting $d_t^{\mathrm{G}} = \|\mathbf{p}_{\mathrm{G}} - \mathbf{p}_t\|_2$ denote the remaining distance to goal, the navigation term $r_t^{\mathrm{nav}}$ combines potential-based progress, flight-time cost, control effort, and terminal arrival bonus with $w_{\mathrm{prog}}, w_t, R_{\mathrm{goal}} > 0$:
\begin{equation}
	\begin{aligned}
		r_t^{\mathrm{nav}} = \; &w_{\mathrm{prog}} \big( d_t^{\mathrm{G}} - d_{t+1}^{\mathrm{G}} \big) - w_t \Delta t - w_a \|\mathbf{a}_t\|_2^2 \\
		&+ R_{\mathrm{goal}} \, \mathbb{I}\big( d_{t+1}^{\mathrm{G}} \le d_{\mathrm{goal}} \wedge d_{\mathrm{obs},t+1} > 0 \big),
	\end{aligned}
	\label{eq:spawc-nav-reward}
\end{equation}
where $\sum_{t=0}^{T-1}(d_t^{\mathrm{G}} - d_{t+1}^{\mathrm{G}}) = d_0^{\mathrm{G}} - d_T^{\mathrm{G}}$ telescopes to a path-independent constant upon arrival, providing dense credit assignment without biasing minimum-time detours.

Moreover, to supply non-vanishing gradients for the binary outage indicator $O_{t+1}$ and enforce obstacle clearance~\eqref{eq:spawc-collision}, we define the smooth SINR reward $r_t^{\mathrm{sinr}} \le 0$ and safety-buffer reward $r_t^{\mathrm{obs}} \le 0$ as
\begin{equation}
	r_t^{\mathrm{sinr}} = \tanh\frac{\gamma_{t+1} - \gamma_{\min}}{\gamma_{\mathrm{margin}}} - 1 - [\gamma_{\min} - \gamma_{t+1}]_+,
	\label{eq:spawc-outage}
\end{equation}
\begin{equation}
	r_t^{\mathrm{obs}} = -\frac{[d_{\mathrm{margin}} - d_{\mathrm{obs},t+1}]_+^2}{d_{\mathrm{margin}}^2} - R_{\mathrm{coll}} \, \mathbb{I}(d_{\mathrm{obs},t+1} \le 0),
	\label{eq:spawc-buffer}
\end{equation}
where $[x]_+ = \max\{0, x\}$, $\gamma_{\mathrm{margin}} > 0$ and $d_{\mathrm{margin}} > 0$ denote the SINR and obstacle buffer margins, respectively, and $R_{\mathrm{coll}} > 0$ is the collision penalty. Conditioning the policy $\pi_{\phi}(\mathbf{a}_t \mid \mathcal{I}_t)$ on the observable history $\mathcal{I}_t = (\mathbf{o}_{0:t}, \mathbf{a}_{0:t-1})$, with $\mathcal{I}_0 = \mathbf{o}_0$, the POMDP objective maximizes expected discounted return:
\begin{equation}
	\max_{\phi} \; J(\phi) = \mathbb{E}_{\pi_{\phi}}\!\left[ \sum_{t=0}^{T-1} \eta^t r_t \right]\!.
	\label{eq:spawc-pomdp-obj}
\end{equation}

\subsection{Multimodal World Model for Belief State Inference}

Directly optimizing~\eqref{eq:spawc-pomdp-obj} over growing histories $\mathcal{I}_t$ is sample-inefficient. Following DreamerV3~\cite{hafner2023dreamerv3}, we encode multimodal observations into compact latent states and predict future latent states, rewards, and continuation flags conditioned on memory and action through an RSSM. Trajectories collected during interaction are stored in a replay buffer $\mathcal{D}$ for off-policy training. Parameterized by $\theta$, the RSSM consists of:
\begin{equation}
	\begin{aligned}
		\text{Sequence model:} \quad &\mathbf{h}_t = f_{\theta}(\mathbf{h}_{t-1}, \mathbf{z}_{t-1}, \mathbf{a}_{t-1}), \\
		\text{Encoder:} \quad &\mathbf{z}_t \sim q_{\theta}(\mathbf{z}_t \mid \mathbf{h}_t, \mathbf{o}_t), \\
		\text{Dynamics predictor:} \quad &\hat{\mathbf{z}}_t \sim p_{\theta}(\hat{\mathbf{z}}_t \mid \mathbf{h}_t), \\
		\text{Reward predictor:} \quad &\hat{r}_t \sim p_{\theta}(\hat{r}_t \mid \mathbf{s}_t), \\
		\text{Continue predictor:} \quad &\hat{\nu}_t \sim p_{\theta}(\hat{\nu}_t \mid \mathbf{s}_t), \\
		\text{Decoder:} \quad &\hat{\mathbf{o}}_t \sim p_{\theta}(\hat{\mathbf{o}}_t \mid \mathbf{s}_t).
	\end{aligned}
	\label{eq:spawc-rssm}
\end{equation}
The multimodal encoder receives both high-dimensional range scans and low-dimensional radio-kinematic vectors: a 2D convolutional neural network (CNN) processes the normalized LiDAR image $\mathbf{L}_t / R_{\mathrm{L}}$, while multilayer perceptrons (MLPs) process the symlog-transformed vectors $\mathbf{s}_t^{\mathrm{radio}}$ and $\mathbf{x}_t^{\mathrm{ego}}$, concatenating their outputs into a fused embedding $\mathbf{e}_t$ before entering the RSSM. Through the recurrent model, observation histories are mapped to a Markovian belief state $\mathbf{s}_t = (\mathbf{h}_t, \mathbf{z}_t)$, featuring a deterministic state $\mathbf{h}_t \in \mathbb{R}^{d_h}$ updated by a block gated recurrent unit (GRU) receiving $\mathbf{a}_{t-1}$ from the policy, and a stochastic state $\mathbf{z}_t \in \{0, 1\}^{K \times C}$ comprising $K$ categorical variables of $C$ classes each. The stochastic state is predicted from $\mathbf{h}_t$ via the prior $p_{\theta}$ during latent imagination, and inferred together with $\mathbf{e}_t$ via the posterior $q_{\theta}$ when live observations are available. From $\mathbf{s}_t$, the decoder reconstructs the multimodal inputs, while the reward and continuation heads employ symexp two-hot categorical and Bernoulli distributions, respectively.

Given batches of $B$ replayed sequences of length $L$, all world model parameters $\theta$ are trained jointly by minimizing
\begin{equation}
	\mathcal{L}_{\mathrm{WM}}(\theta) = \mathbb{E}_{q_{\theta}}\!\Big[ \sum_{t=1}^{L} \!\big( \beta_{\mathrm{pred}} \mathcal{L}_t^{\mathrm{pred}} + \beta_{\mathrm{dyn}} \mathcal{L}_t^{\mathrm{dyn}} + \beta_{\mathrm{rep}} \mathcal{L}_t^{\mathrm{rep}} \big) \Big],
	\label{eq:spawc-wm-loss}
\end{equation}
with prediction, dynamics, and representation losses defined as
\begin{equation}
	\begin{aligned}
		\mathcal{L}_t^{\mathrm{pred}} &= -\ln p_{\theta}(\mathbf{o}_t \mid \mathbf{s}_t) - \ln p_{\theta}(r_t \mid \mathbf{s}_t) - \ln p_{\theta}(\nu_t \mid \mathbf{s}_t), \\
		\mathcal{L}_t^{\mathrm{dyn}} &= \max\!\big(1, \, D_{\mathrm{KL}}\big[\mathrm{sg}(q_{\theta}(\mathbf{z}_t \mid \mathbf{h}_t, \mathbf{o}_t)) \,\|\, p_{\theta}(\hat{\mathbf{z}}_t \mid \mathbf{h}_t)\big]\big), \\
		\mathcal{L}_t^{\mathrm{rep}} &= \max\!\big(1, \, D_{\mathrm{KL}}\big[q_{\theta}(\mathbf{z}_t \mid \mathbf{h}_t, \mathbf{o}_t) \,\|\, \mathrm{sg}(p_{\theta}(\hat{\mathbf{z}}_t \mid \mathbf{h}_t))\big]\big),
	\end{aligned}
	\label{eq:spawc-kl}
\end{equation}
where $\mathrm{sg}(\cdot)$ stops gradients and $1\,\mathrm{nat}$ free bits prevent posterior collapse.

\subsection{Policy Optimization in Latent Imagination}

To solve~\eqref{eq:spawc-pomdp-obj} without extra environment interactions, we freeze $\theta$ and train a stochastic actor $\tilde{\mathbf{a}}_\tau \sim \pi_{\phi}(\tilde{\mathbf{a}}_\tau \mid \tilde{\mathbf{s}}_\tau)$ with policy mean $\boldsymbol{\mu}_\tau = \mathbb{E}[\pi_{\phi}(\cdot \mid \tilde{\mathbf{s}}_\tau)]$ and a distributional critic $v_{\psi}(G_\tau^{\lambda} \mid \tilde{\mathbf{s}}_\tau)$ with expected value $V_{\psi}(\tilde{\mathbf{s}}_\tau) = \mathbb{E}[v_{\psi}(\cdot \mid \tilde{\mathbf{s}}_\tau)]$ purely on imagined latent trajectories~\cite{hafner2023dreamerv3}. Starting from detached posteriors $\tilde{\mathbf{s}}_t = \mathrm{sg}(\mathbf{s}_t)$ along replayed sequences, $f_{\theta}$ and $p_{\theta}(\hat{\mathbf{z}} \mid \mathbf{h})$ unroll imagined rollouts $\{\tilde{\mathbf{s}}_\tau, \tilde{\mathbf{a}}_\tau, \hat{r}_\tau, \hat{\nu}_{\tau+1}\}_{\tau=t}^{t+H-1}$ over horizon $H$ under $\|\tilde{\mathbf{a}}_\tau\|_2 \le a_{\max}$ without decoding observations. Reading out $\hat{r}_\tau$ and $\hat{\nu}_{\tau+1}$, bootstrapped $\lambda$-returns with $\lambda \in [0, 1]$ are computed recursively for $\tau = t+H-1, \dots, t$ to capture rewards beyond horizon $H$:
\begin{equation}
	\begin{aligned}
		G_\tau^{\lambda} &= \hat{r}_\tau + \eta \hat{\nu}_{\tau+1}\Big[(1 - \lambda)V_{\psi}(\tilde{\mathbf{s}}_{\tau+1}) + \lambda G_{\tau+1}^{\lambda}\Big], \\
		G_{t+H}^{\lambda} &= V_{\psi}(\tilde{\mathbf{s}}_{t+H}).
	\end{aligned}
	\label{eq:spawc-td}
\end{equation}
The critic $v_{\psi}$ is trained to predict $\mathrm{sg}(G_\tau^{\lambda})$ by minimizing the symexp two-hot negative log-likelihood $-\ln v_{\psi}(\mathrm{sg}(G_\tau^{\lambda}) \mid \tilde{\mathbf{s}}_\tau)$ stabilized by an exponential moving average (EMA) target network.

Letting $\tilde{w}_\tau = \mathrm{sg}(\prod_{k=t+1}^{\tau} \eta \hat{\nu}_k)$ with $\tilde{w}_t = 1$ denote cumulative discount weights and $S$ denote the EMA of the $5\%$--$95\%$ percentile range of $G_\tau^{\lambda}$ across the batch, the actor $\pi_{\phi}$ is trained via return-normalized REINFORCE with entropy $\mathcal{H}[\pi_{\phi}]$ weighted by $\xi > 0$ and action-mean smoothness weighted by $w_{\mathrm{smooth}} \ge 0$:
\begin{equation}
	\begin{aligned}
		\mathcal{L}_{\mathrm{act}}(\phi) ={}& -\frac{1}{H}\sum_{\tau=t}^{t+H-1} \mathbb{E}\!\Bigg[ \tilde{w}_\tau \Bigg( \mathrm{sg}\!\left(\frac{G_\tau^{\lambda} - V_{\psi}(\tilde{\mathbf{s}}_\tau)}{\max(1, S)}\right) \\
		&\quad \times \ln\pi_{\phi}(\tilde{\mathbf{a}}_\tau \mid \tilde{\mathbf{s}}_\tau) + \xi \mathcal{H}\big[\pi_{\phi}(\cdot \mid \tilde{\mathbf{s}}_\tau)\big] \Bigg) \Bigg] \\
		& + \frac{w_{\mathrm{smooth}}}{H-1}\sum_{\tau=t}^{t+H-2} \mathbb{E}\!\left[ \|\boldsymbol{\mu}_{\tau+1} - \boldsymbol{\mu}_\tau\|_2^2 \right]\!.
	\end{aligned}
	\label{eq:spawc-actor}
\end{equation}
Applying smoothness to $\boldsymbol{\mu}_\tau$ rather than sampled stochastic actions preserves exploration variance during training while ensuring jerk-free flight commands during execution. At online deployment, all decoders, reward and continue heads, and the critic $v_{\psi}$ are discarded. At each slot $t$, only the multimodal encoder, GRU $f_{\theta}$, posterior $q_{\theta}(\mathbf{z}_t \mid \mathbf{h}_t, \mathbf{o}_t)$, and deterministic policy mean $\mathbf{a}_t = a_{\max}\boldsymbol{\mu}_t / \max\{1, \|\boldsymbol{\mu}_t\|_2\}$ execute onboard without online search or maps.

\section{Numerical Results}

\begin{table}[!t]
	\centering
	\caption{System, 3GPP UMa Radio, and World Model Parameters}
	\label{tab:spawc-params}
	\renewcommand{\arraystretch}{1.03}
	\resizebox{\columnwidth}{!}{%
	\begin{tabular}{ll}
		\toprule
		\textbf{Parameter} & \textbf{Value} \\
		\midrule
		Urban Domain $\mathcal{W}$ \& City Split & $1000\,\mathrm{m} \times 1000\,\mathrm{m}$ ($8$ Train / $2$ Test Cities) \\
		BS Deployment \& Array ($M$) & $5\,\mathrm{sites/km^2}$, $3$ Sectors/Site ($M = 15$), $8 \times 8$ \\
		BS Height $h_{\mathrm{BS}}$ \& Downtilt & $25.0\,\mathrm{m}$ AGL, $8.0^\circ$ ($0^\circ, 120^\circ, 240^\circ$) \\
		Carrier $f_c$ \& Bandwidth $W$ & $4.9\,\mathrm{GHz}$ (5G n79), $100\,\mathrm{MHz}$ \\
		Powers ($P_{\mathrm{tx},m}, P_{\mathrm{ref}}$) \& Noise & $49.0\,\mathrm{dBm}$, $1\,\mathrm{mW}$ ($\mathrm{NF} = 9.0\,\mathrm{dB}$) \\
		Traffic Load $\rho_{\mathrm{load}}$ \& Threshold $\gamma_{\min}$ & $\rho_{\mathrm{load}} = 0.75$, $\gamma_{\min} = 0.0\,\mathrm{dB}$ \\
		Ray-Tracing Depth \& Cell Reports $M_{\mathrm{obs}}$ & Depth $5$ ($2 \times 10^6$ Rays/Sector), $M_{\mathrm{obs}} = 6$ \\
		\midrule
		Cruise Altitude $H_{\mathrm{uav}}$ \& Slot $\Delta t$ & $65.0\,\mathrm{m}$, $\Delta t = 0.2\,\mathrm{s}$ ($5\,\mathrm{Hz}$ / $20\,\mathrm{Hz}$) \\
		Kinematic Bounds ($v_{\max}, a_{\max}$) & $15.0\,\mathrm{m/s}$, $12.0\,\mathrm{m/s^2}$ \\
		Radii ($d_{\mathrm{goal}}, d_{\mathrm{margin}}$) & $20.0\,\mathrm{m}$, $22.0\,\mathrm{m}$ \\
		Spherical 3D LiDAR ($N_{\mathrm{el}} \times N_{\mathrm{az}}, R_{\mathrm{L}}$) & $16 \times 64$ ($[-30^\circ, 30^\circ]$), $R_{\mathrm{L}} = 100\,\mathrm{m}$ \\
		\midrule
		RL Episode Step Limit $T_{\max}$ & $T_{\max} = 750$ Steps ($150\,\mathrm{s}$) \\
		Navigation Reward ($w_{\mathrm{prog}}, w_t, w_a, R_{\mathrm{goal}}$) & $0.60$, $0.25$, $5 \times 10^{-4}$, $100.0$ \\
		SINR Reward ($w_{\mathrm{sinr}}, \gamma_{\mathrm{margin}}$) & $1.75$, $4.0\,\mathrm{dB}$ \\
		Obstacle Reward ($w_{\mathrm{obs}}, R_{\mathrm{coll}}$) & $2.50$, $100.0$ \\
		Categorical RSSM ($d_h, K \times C$) & $d_h = 256$, $32 \times 32$ ($1\%$ Unimix) \\
		Batch, Sequence, \& Horizon ($B, L, H$) & $B = 16$, $L = 60$, $H = 38$ \\
		WM Loss Scales ($\beta_{\mathrm{pred}}, \beta_{\mathrm{dyn}}, \beta_{\mathrm{rep}}$) & $1.0$, $1.0$, $0.1$ ($1\,\mathrm{nat}$ Free Bits) \\
		Symexp Two-Hot Distribution & $255$ Bins over $[-20, 20]$ \\
		Discount \& $\lambda$-Return ($\eta, \lambda$) & $\eta = 0.996$, $\lambda = 0.98$ ($5\%\text{--}95\%$ Return Scale) \\
		Entropy \& Smoothness ($\xi, w_{\mathrm{smooth}}$) & $\xi = 3 \times 10^{-4}$, $w_{\mathrm{smooth}} = 0.05$ \\
		\bottomrule
	\end{tabular}%
	}
	\vspace{-0.18cm}
\end{table}

\begin{table}[!t]
	\centering
	\caption{Zero-Shot Performance Across Unseen Urban Test Cities}
	\label{tab:spawc-results}
	\renewcommand{\arraystretch}{1.03}
	\resizebox{\columnwidth}{!}{%
	\begin{tabular}{lccccccc}
		\toprule
		\textbf{Planner} & \textbf{Success} & \textbf{Collision} & \textbf{Eff.\ Outage} & \textbf{Route SINR} & \textbf{Route Outage} & \textbf{Steps} & \textbf{Return} \\
		\midrule
		\textbf{Proposed CommDreamer} & $\mathbf{100.0\%}$ & $\mathbf{0.0\%}$ & $\mathbf{12.0\%}$ & $\mathbf{7.69\,\mathrm{dB}}$ & $\mathbf{12.0\%}$ & $325.7$ & $\mathbf{422.80}$ \\
		\textbf{Geometry-Only WM} & $93.8\%$ & $6.2\%$ & $24.2\%$ & $5.91\,\mathrm{dB}$ & $19.1\%$ & $\mathbf{316.3}$ & $355.40$ \\
		\textbf{Reactive LiDAR APF} & $87.5\%$ & $12.5\%$ & $24.8\%$ & $6.10\,\mathrm{dB}$ & $14.1\%$ & $365.5$ & $332.22$ \\
		\textbf{Model-Free PPO} & $0.0\%$ & $6.2\%$ & $100.0\%$ & --$^*$ & --$^*$ & $708.8$ & $-113.16$ \\
		\bottomrule
		\multicolumn{8}{l}{\footnotesize $^*$Evaluated over completed flights; failed runs count as $100\%$ in Eff.\ Outage.}
	\end{tabular}%
	}
	\vspace{-0.18cm}
\end{table}

\begin{figure}[!t]
	\centering
	\includegraphics[width=0.88\columnwidth]{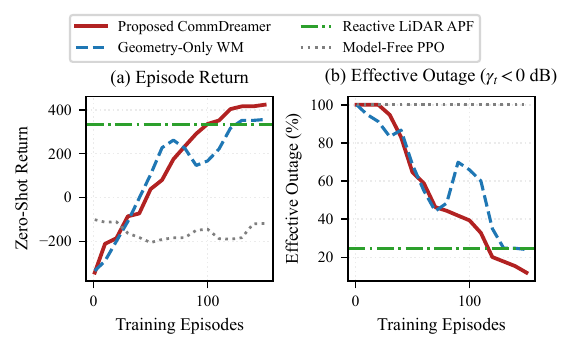}
	\caption{Zero-shot evaluation curves across unseen test cities during training: (a) episode return and (b) effective SINR outage ratio under $\gamma_t < 0\,\mathrm{dB}$.}
	\label{fig:spawc-conv}
	\vspace{-0.20cm}
\end{figure}

\begin{figure}[!t]
	\centering
	\includegraphics[width=0.92\columnwidth]{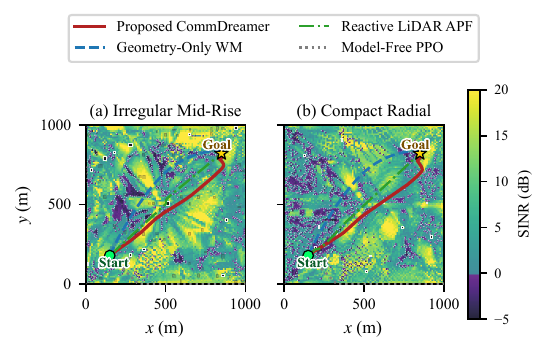}
	\caption{Zero-shot closed-loop trajectories over the hidden Sionna RT SINR maps in unseen test cities: (a) \textit{Irregular Mid-Rise} and (b) \textit{Compact Radial Low-Rise}.}
	\label{fig:spawc-traj}
	\vspace{-0.18cm}
\end{figure}

\begin{figure}[!t]
	\centering
	\includegraphics[width=0.88\columnwidth]{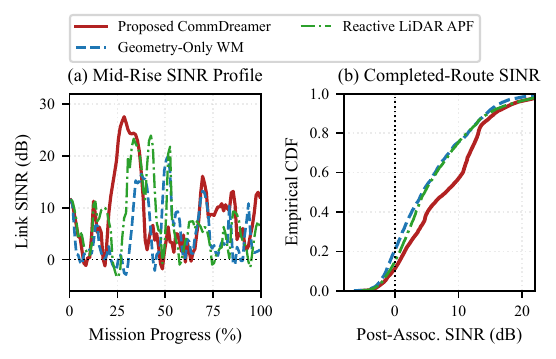}
	\caption{Link-quality comparison across completed flights: (a) representative \textit{Irregular Mid-Rise} SINR profile along mission progress and (b) empirical CDF of post-association SINR over all completed test missions.}
	\label{fig:spawc-sinr}
	\vspace{-0.18cm}
\end{figure}

\subsection{Simulation Setup and Baselines}

We evaluate \textit{CommDreamer} across procedural 3D urban canyons built with our lightweight open-source simulator \textit{AeroCity-RT}, which integrates deterministic 3D city generation and 3GPP BS configuration with site-specific electromagnetic ray tracing in Sionna RT~\cite{hoydis2023sionna_rt}, with key parameters listed in Table~\ref{tab:spawc-params}. Each city deploys tri-sector 3GPP UMa macro sites equipped with dual-polarized directional planar arrays over ITU concrete buildings and dry ground terrain, tracing LoS transmission, specular reflection, and edge diffraction. Policies are trained for $150$ episodes across $8$ procedural training cities with randomized cross-city start-destination pairs, and are evaluated zero-shot without fine-tuning across $8$ cross-city missions in each of the unseen \textit{Irregular Mid-Rise} and \textit{Compact Radial Low-Rise} test cities. Under an identical sample budget, we benchmark \textit{CommDreamer} against three representative baselines: \textit{Geometry-Only WM}, an ablation sharing the 3D LiDAR CNN, categorical RSSM, and $H$-step latent imagination while setting $\mathbf{s}_t^{\mathrm{radio}} = \mathbf{0}$ and $w_{\mathrm{sinr}} = 0$; \textit{Reactive LiDAR APF}, an artificial potential field planner driven by goal attraction and 3D LiDAR obstacle repulsion; and \textit{Model-Free PPO}, an on-policy actor-critic sharing the multimodal encoder in Section~III-B.

\subsection{Sample Efficiency and Zero-Shot Generalization}

As shown in Fig.~\ref{fig:spawc-conv} and Table~\ref{tab:spawc-results}, unrolling $H = 38$ imagined steps over a $7.6\,\mathrm{s}$ horizon in the categorical RSSM enables \textit{CommDreamer} to converge within $150$ episodes, achieving $100.0\%$ zero-shot mission success with zero collisions, $7.69\,\mathrm{dB}$ completed-route SINR, and $12.0\%$ route and effective SINR outage across unseen cities. Conversely, \textit{Model-Free PPO} fails to complete cross-city flights under partial observability and sparse arrival bonuses, yielding $0.0\%$ success and $6.2\%$ collisions, whereas \textit{Reactive LiDAR APF} suffers local-minima collisions in dense non-convex canyons with $87.5\%$ overall success, $12.5\%$ collisions overall and $25.0\%$ in \textit{Irregular Mid-Rise}, $6.10\,\mathrm{dB}$ route SINR, $14.1\%$ route outage, and $24.8\%$ effective outage.

Compared with radio-blind \textit{Geometry-Only WM}, which records $93.8\%$ success, $6.2\%$ collisions overall and $12.5\%$ in \textit{Irregular Mid-Rise}, $5.91\,\mathrm{dB}$ route SINR, $19.1\%$ route outage, and $24.2\%$ effective outage, \textit{CommDreamer} takes only $9.4$ extra steps or $1.9\,\mathrm{s}$ to detour around deep shadow zones, eliminating collisions while cutting route outage by $37.0\%$ relative and reducing effective outage by $50.1\%$ and $51.5\%$ relative to \textit{Geometry-Only WM} and \textit{Reactive LiDAR APF}. As illustrated in Fig.~\ref{fig:spawc-traj}, \textit{Geometry-Only WM} cuts directly across central NLoS interference zones with $4.9\,\mathrm{dB}$ mean SINR and $15.3\%$ outage in \textit{Irregular Mid-Rise} and $4.2\,\mathrm{dB}$ mean SINR and $29.3\%$ outage in \textit{Compact Radial Low-Rise}, whereas \textit{CommDreamer} anticipates corner blockages and tracks high-gain LoS corridors at $10.6\,\mathrm{dB}$ with $3.7\%$ outage and $9.4\,\mathrm{dB}$ with $6.6\%$ outage, respectively. Consistently across all completed test missions in Fig.~\ref{fig:spawc-sinr}, \textit{CommDreamer} shifts the empirical SINR CDF substantially rightward, lifting median SINR from $4.38\,\mathrm{dB}$ under \textit{Geometry-Only WM} and $4.71\,\mathrm{dB}$ under \textit{Reactive LiDAR APF} to $8.02\,\mathrm{dB}$ while boosting overall route SINR by $1.59\text{--}1.78\,\mathrm{dB}$.

\section{Conclusion}

We presented \textit{CommDreamer}, a mapless cellular-connected UAV trajectory planner built on a canonical DreamerV3 radio-geometric world model. By fusing spherical 3D LiDAR scans, multi-cell RSRP/SINR reports, and relative kinematics into a categorical RSSM belief state and training an end-to-end policy via single-critic latent imagination and return-normalized REINFORCE, the UAV proactively avoids NLoS outage pockets in unseen cities without prior maps, channel priors, or heuristic rules. Future work will explore primal-dual hard safety constraints and 3D altitude maneuvers.

\def\BIBdecl{\let\BIBentrySTDinterwordspacing\BIBentryALTinterwordspacing}
\bibliographystyle{IEEEtran}
\bibliography{reference}

\end{document}